# CAUSALGRAPH: A PYTHON PACKAGE FOR MODELING, PERSISTING AND VISUALIZING CAUSAL GRAPHS EMBEDDED IN KNOWLEDGE GRAPHS

Sven Pieper [a,*], Carl Willy Mehling [a,*], Dominik Hirsch [a],
Tobias Lüke [a] and Steffen Ihlenfeldt [a,b]

[a] Fraunhofer Institute for Machine Tools and Forming Technology IWU, 01187 Dresden, Germany
[b] Technische Universität Dresden, Institute of Mechatronic Engineering, 01062 Dresden, Germany
* Correspondence: {sven.pieper, carl.willy.mehling}@iwu.fraunhofer.de

## ABSTRACT

This paper describes a novel Python package, named `causalgraph`, for modeling and saving causal graphs embedded in knowledge graphs. The package has been designed to provide an interface between causal disciplines such as causal discovery and causal inference. With this package, users can create and save causal graphs and export the generated graphs for use in other graph-based packages. The main advantage of the proposed package is its ability to facilitate the linking of additional information and metadata to causal structures. In addition, the package offers a variety of functions for graph modeling and plotting, such as editing, adding, and deleting nodes and edges. It is also compatible with widely used graph data science libraries such as `NetworkX` and `Tigramite` and incorporates a specially developed `causalgraph` ontology in the background. This paper provides an overview of the package's main features, functionality, and usage examples, enabling the reader to use the package effectively in practice.

The package `causalgraph` is available at `https://github.com/causalgraph/causalgraph`.

***K**eywords* Causality · Knowledge Graph · RDF · Causal Inference · Causal Discovery · NetworkX · Tigramite · Python

## 1 Introduction

The application of causal analysis has been a significant driver in the recent advances in analytics and artificial intelligence. Structuring a problem in cause-effect pairs makes it possible to accurately predict the results of particular interventions on any given phenomenon. This analysis enables one to answer causal questions without resorting to randomized control trials and is invaluable in various scientific and business endeavors.

Two pivotal elements that enabled the upsurge of causality-driven analysis were the graphical representation of causality in Directed Acyclic Graphs (DAGs) and the mathematical formalization of interventions, known as "do-calculus" [1]. These two critical components have been instrumental in advancing research in this field, continuously improved upon by causal inference and causal discovery disciplines.

The foundation of causal inference (CI) consists of a Directed Acyclic Graph and data to quantify interventions. The DAG encodes causal relations and, more crucially, the causal independencies of the issue at hand. The data is then used to approximate the causal links through the application of models, which can either be probabilistic (Causal Bayesian Networks) or deterministic (Structural Causal Models). Thus, the combination of the DAG and models constitutes a Causal Graphic Model (CGM), which is used to quantify the effect of interventions on a given state of the graph. While the models are fitted with data, the DAG is usually established from an expert's knowledge through a manual process. This task can be increasingly arduous with the rise in the number of variables, and often, the causal structure of a problem remains unknown. To provide solutions to such issues, causal discovery methods have been developed.

Causal discovery (CD) tries to construct DAGs from observations. Hence, causal dependencies need to be identified in the data and oriented in the direction from cause to effect. CD is an inverse problem, and the reconstruction of the causal structure from observations is not uniquely solvable. The underlying causal mechanism can only be discovered until the so-called Markov equivalence class [2]. Especially for many variables and time-series data, this reduction of possible DAGs is essential and improved algorithms continuously increase performance for causal discovery.

Although both CI and CD revolve around the concept of causal graphs, no standardized way of representing causal information exists in this graph format. This lack of a unified representation of causal information leads to difficulties when attempting to combine CD and CI, as the implementations utilized in each field are typically distinctive, with no established exchange format available to bridge the gap.

Various implementations for the representation of simple DAGs exist, but these need to be augmented with additional information to be employed as a full container for a causal graph. Inherently, a DAG encodes the causal structure through its nodes and edges. Nevertheless, models and context information that describe these nodes and edges need to be appended. Notably, in the case of causal discovery, context information regarding the edges is indispensable, such as the algorithm's conviction in the edge's presence and the time lag between cause and effect for temporal data. Usually, this information is stored in matrices to allow for efficient computation, but it is lost when transferred to a basic graph notation.

The development of a Python package is intended to prevent the loss of information and promote the seamless transfer between CD and CI. This package stores the causal structure and complementary data within a knowledge graph. This idea was suggested by Jaimini et al. in [3], which highlighted the advantages of explainability regarding the results of CI. Our work employs a less complex ontology, sufficient to create a continuous, context-rich description of causal graphs and be compatible with existing CD and CI tools. The proposed Python package `causalgraph` furnishes modeling and plotting functions for causal graphs and a storage database that allows for an ongoing adaptation of the causal information at runtime. Additionally, imported and exported capabilities from various tools are available, enabling the capacity to integrate a broad array of CI and CD implementations effectively.

## 2 Building Blocks

To gain a deeper understanding of the capabilities of the `causalgraph` package, it is crucial to become familiar with its main components and functionalities. Of particular note is the `causalgraph` ontology, which was created explicitly for this package, the integration of Resource Description Framework (RDF) into `causalgraph`, and the notion of RDF Reification. Each of these elements is fundamental for effective software utilization and will be expounded upon in this chapter.

### 2.1 RDF in causalgraph

The World Wide Web Consortium's (W3C) Resource Description Framework (RDF) has received increased attention in recent years as a data representation format for Linked Data, knowledge graphs, and the Semantic Web [4][5]. Due to RDF providing an effective means for integrating data from multiple sources, interlinking them, and efficiently querying them with query languages such as SPARQL Protocol and RDF Query Language (SPARQL) [5][6][7]. This makes RDF pivotal in constructing graph-based data and a fundamental building block for describing and exchanging information.

The Resource Description Framework (RDF) is a standardized data interchange format that enables the representation of highly interconnected data [8]. It captures the binary entities that link two different objects and expresses facts, relationships, and data through its direct connections between resources. This is done by structuring each statement in its uniform form as a triple, consisting of three different resources identifiable by their own Uniform Resource Identifier (URI). To illustrate, a triple could look like this: *"Berlin hasFlightConnectionTo NewYork"*. This structure allows for data to be structured and organized orderly while also enabling connections between other triple components, thereby resulting in a highly interconnected data structure [9].

The generation of knowledge graphs through `causalgraph` is ultimately the result of highly interconnected information in the form of causal relationships. As a result, RDF stands out as a practical basis for storing causal relationships. To this end, the Python package `owlready2` [10] has been incorporated into `causalgraph`, allowing for the abstraction of the RDF information representation and the direct integration and manipulation of the required ontologies. This is advantageous as it allows for the instantiation of graph individuals without concrete RDF knowledge, similar to the commonly used Python package for complex networks called `NetworkX` [11]. Moreover, thanks to `owlready2`, these individuals are generated according to the imported ontologies, thus conducting validity tests for classes, property assignments, names, and target pairs. `Causalgraph` utilizes an `owlready2` triplestore to keep these causal graphs, allowing triples to be stored in the form of subject-predicate-object pacts. The `owlready2` triplestore facilitates the

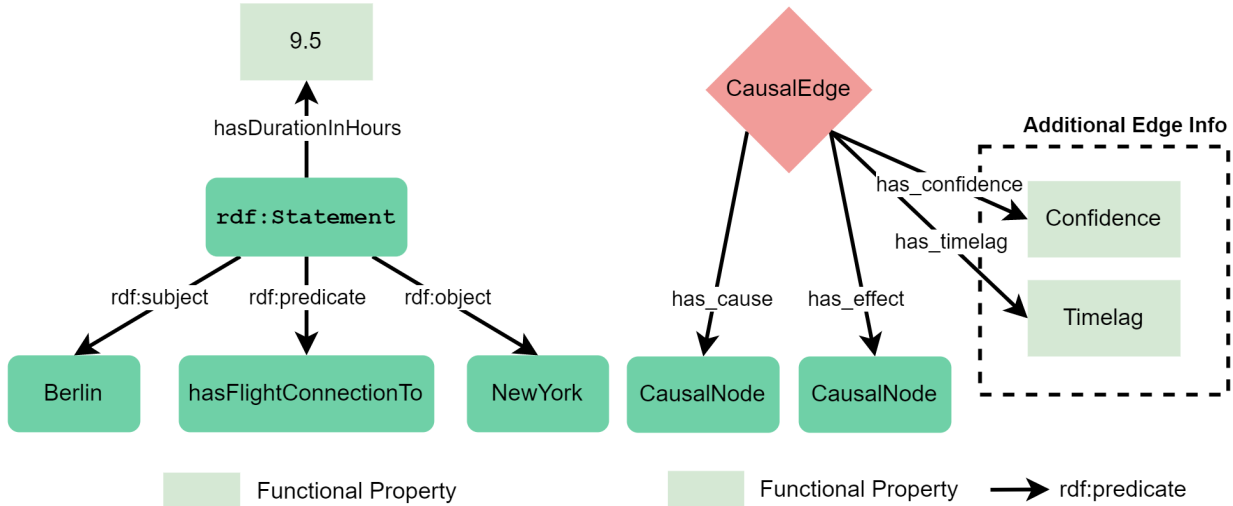

(a) Illustration of RDF reification methodology. The triple "Berlin hasFlightConnectionTo NewYork" is encapsulated in an rdf:Statement, giving it its own URI and enabling the addition of additional metadata such as "hasDuration".

(b) Illustration of RDF reification in causalgraph. The causal relationship is depicted through the use of a CausalEdge connecting CausalNodes. This edge is created as an independent object with its own URI, enabling the addition of metadata such as confidence or time lags.

Figure 1: Illustration of RDF reification methodology and its application in causalgraph. The figures show the encapsulation of a triple in an rdf:Statement and the creation of a CausalEdge as independent object with its own URIs to append metadata.

storage of graphs in an SQL database, which delivers data consistency and continuous operation with `causalgraph`. In addition, native support for optimized SPARQL queries is provided through the use of `owlready2` in `causalgraph`. SPARQL queries not only enable the matching of RDF triples but also grant the ability to employ mathematical operations and other utility functions to quickly filter data [7]. Such queries are also employed in `causalgraph` to extract causal information for the purpose of importing and exporting causal graphs. Furthermore, extra information which is not part of the causal graph itself can be affixed and linked, such as comments assigned to a causal link or the association of a machine component. RDF plays an essential part in `causalgraph`. However, some particular features are still worth mentioning, which will be discussed in the subsequent subchapter.

## 2.2 RDF Reification with causalgraph

As described in the previous subsection, in RDF, knowledge is expressed and represented by triples. Unfortunately, with the standard RDF schema (subject–predicate–>object), it is not possible to add further information to stand-alone triples, so enriching RDF statements with metadata is not feasible without additional effort [8][9]. For `causalgraph`, however, especially with regard to the causal methods mentioned in chapter 1, such as causal inference and causal discovery, it is advantageous to link causal relations (causalgraph RDF triples) to functional values and to persist these within the knowledge graph. This may include, for instance, the time lag of a particular cause-effect mechanism, or the degree of confidence regarding a calculated causal relationship. To further explain this limitation, a brief example can be provided:

*"Berlin has a flight connection to New York, which takes 9.5 hours."*

This statement can be divided into two parts of knowledge: First, the statement that Berlin has a flight connection to New York (`Berlin hasFlightConnectionTo NewYork`) and second, that this flight takes 9.5 hours (`??  hasDuration 9.5h`). To combine both statements, "`[Berlin hasFlightConnectionTo NewYork] hasDuration 9.5h`" would be conceivable, which would require a conversion of the triple "`Berlin hasFlightConnectionTo NewYork`" into a subject in order to comply with the RDF requirements. Regrettably, the RDF standard does not permit the conversion of triples to objects or subjects, since they would have to be assigned a specific Uniform Resource Identifier (URI) for the aforementioned combined statement to be valid.

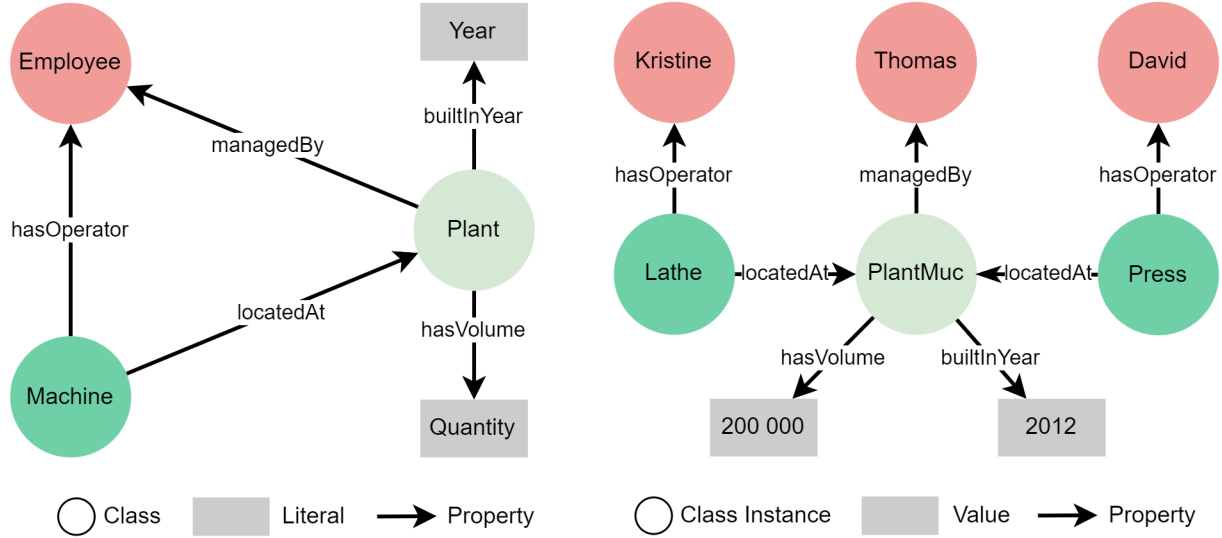

(a) Illustration of an exemplary ontology. The ontology describes the taxonomic-formal knowledge of a plant and its associated machines and responsible persons. This ontology can be used to create a knowledge graph to represent concrete works and their attributes.

(b) Illustration of a generated knowledge graph based on the ontology from figure 2a. The specific concepts and relationships within the domain have been applied to an actual plant, creating multiple instances resulting in the construction of a knowledge graph.

Figure 2: Illustration of an ontology and its corresponding knowledge graph to show the differences between both concepts. Subfigure 2a shows an ontology describing the taxonomic-formal knowledge of plants, and subfigure 2b depicts the resulting knowledge graph when applying the ontology to a specific plant.

Jaimini et al. presented a novel solution to this problem in [3] by leveraging the RDF extension RDF* (pronounced RDF-star) to allow for embedded triples, wherein triples can be made subjects or objects of statements and thus receive their own, independent Uniform Resource Identifier [12]. Unfortunately, the `owlready2` library does not include RDF*, thus necessitating the standard RDF reification instead. Although this approach provides a much less efficient serialization syntax, it still permits the same basic functionality [8]. To do this, an RDF statement (rdf:Statement) is specified, which acts like a container for the triple and thus consists of the three parts: subject, predicate, and object. Utilizing this RDF statement, which has an independent URI, it is possible to create a subject that comprises multiple statements, all of which continue to adhere to the mandatory RDF triple form [8]. This principle is illustrated in figure 1a, wherein the triple "`Berlin hasFlightConnectionTo NewYork`" is associated with the type rdf:Statement, thus resulting in the generation of a URI. This statement can then be used as the subject of another statement to provide additional data, such as the flight duration of this particular connection. Thus, a triple has been extended by further metadata.

The principles of reification outlined herein have been adapted from `causalgraph`, depicted in figure 1b. This model illustrates the causal relationship between a cause and an effect by utilizing a designated edge, known as a CausalEdge, between the two nodes, referred to as CausalNodes. This CausalEdge, which connects the cause and effect, is created as an independent RDF object with its own URI. This enables the system to be further enriched with additional metadata, such as a measure of confidence in the connection or a time lag, in line with the RDF reification described above.

### 2.3 Causalgraph Ontology

The construction of knowledge graphs that capture the knowledge from various sources and demonstrate the connections between the individual components requires using an ontology [13][14]. This is due to the ontology providing a taxonomic-formal description of the fundamental knowledge regarding concepts and relationships within a specific domain [15]. Once the ontology has been established, it serves as the data model for creating a knowledge graph, with concrete data structured in accordance with the ontology [13]. Figure 2 demonstrates this distinction between the two concepts.

In order to effectively systematize knowledge, concepts, and relationships, ontologies must incorporate a variety of different components. This includes entities such as classes, properties, and individuals. Moreover, it involves

CAUSALGRAPH: CAUSAL GRAPHS EMBEDDED IN KNOWLEDGE GRAPHSthe establishment of statements to describe class restrictions and axioms to more accurately define subclasses [16]. Ultimately, careful curation of such a knowledge model in a given domain can aid in generating and sharing a deeper understanding of the subject. Through this process, the information contained in the ontology can be further extended, as new relationships and concepts can be easily added while existing ones can be updated or revised [17]. Furthermore, the knowledge graph can be enriched with metadata to enhance data quality or trace data origins. As a result, the ontology can expand and grow as the data volume increases without affecting dependent processes or systems [17].

Ontologies are of central importance for applying the package `causalgraph`, as `causalgraph` allows for integrating causal graphs into knowledge graphs. As such, an ontology can offer a formal reference point for causal concepts and the addition of metadata. To this end, a special `causalgraph` ontology has been developed and released alongside the Python package[1]. The two primary entities of this ontology are CausalNodes and CausalEdges. When two CausalNodes are linked causally, an edge between them is drawn and mapped as a CausalEdge. The object properties "isAffectedBy" and "isCausing" can be assigned to the CausalNodes to register the link to a specific CausalEdge, as well as the object properties "hasEffect" and "hasCause" to the CausalEdge, with the involved CausalNodes specified. Furthermore, the data type properties "hasConfidence" and "hasTimeLag" can be utilized to enrich the relationship between the CausalNodes, as can be seen in figure 1b. Additionally, an exclusive Creator class is defined in the ontology to characterize the source of particular causal connections in the knowledge graph. This creation of causal relations could result from an expert-driven, manual process or a software-based scheme such as causal discovery. The object properties "hasCreator" and "created" can be used to store this provenance information in the knowledge graph. Moreover, the `causalgraph` ontology presents subclasses of CausalNodes, including States, Events, and Variables, which add more descriptive details about the CausalNode. States stands for a type of signal that can switch between different states, whereas a Variable symbolizes a continuous signal, thus not being able to be a state or an event. Events, in contrast, refer to event-like signals, such as errors or warnings in a system.

## 3 Functionalities

To use `causalgraph`, the package must first be installed. This can be done via the package installer Pip [18], which also installs all required packages. The current source code and documentation are also available via the GitHub repository[2]. After the installation, it is possible to import `causalgraph` into Python and use its functions. The handling is inspired by the widely used graph package `NetworkX` [11]. One of the main functions of `causalgraph` is the modeling of graphs. For this purpose, nodes and edges can be added, removed, or edited. The package also offers the possibility to import and export graphs and thus use them in other graph-based packages such as `Tigramite` [19]. Furthermore, it is possible to display the modeled graphs graphically using the visualization functions provided. For using knowledge and concepts from other ontologies, `causalgraph` allows importing external ontologies. The starting point for modeling a graph is a `Graph()` object. The instantiation of an empty graph is shown in listing 1. The graph is stored in a .sqlite-file, and the location can be defined via the optional `sql_db_filename` parameter. Parallel reading and writing of the database can be controlled via the `sql_exclusive` parameter. The location of the logging file is defined via `log_file_dir`. To load third-party ontologies during instantiation, it is possible to pass a list of ontology paths via the parameter `external_ontos`.

Listing 1: The following code demonstrates how to create and initialize a `causalgraph` `Graph()` object named `G` with optional parameters. The parameters include options for SQL database access regulations, log file, logger verbosity, external ontologies, and external graph input.

```
>>> from causalgraph import Graph
>>> G = Graph(
...     # Optional
...     sql_db_filename="<path_to_sql_file>",
...     sql_exclusive=False,
...     log_file_dir="<path_to_log_dir>",
...     logger_level=10,
...     external_ontos=["<path_to_onto>", "<path_to_onto>"],
...     external_graph="<nx.MultiDiGraph | Tigramite Tuple>")
```

---

[1]The causalgraph ontology and documentation available at https://github.com/causalgraph/causalgraph-ontology
[2]The causalgraph source code and documentation available at https://github.com/causalgraph/causalgraph/

## 3.1 Modeling

In principle, two essential graph variables are available for modeling graphs. A distinction must be made between `add` for inserting edges and `remove` for deleting them. The creation of new nodes in the graph is done using the `add` function `causal_node()`, whose usage is shown in listing 2. When naming the nodes, it is essential to say that only strings can be passed. In case a node with the desired name already exists, no additional node will be created.

> Listing 2: The `Graph()` object permits the addition of `CausalNodes` through the use of the `add.causal_node()` function. It is necessary to designate a name for the node, and should this not be done, a generic name is assigned by default. Furthermore, it is possible to include other relevant information such as comments. Upon successful creation of the node, the name of the node is returned.

```
>>> G.add.causal_node(individual_name="Rain")
>>> node_name = G.add.causal_node(individual_name="Wet", comment=["some text"])
```

Adding edges is done similarly except that the `causal_edge()` function must be used for this, as can be seen in listing 3. The three passing parameters `cause_node_name`, `effect_node_name` and `name_for_edge` are crucial here. Setting `force_create=True` ensures that causal nodes that do not yet exist are instantiated when the edge is created. When the CausalEdge is created, the two properties `hasCause` and `hasEffect` will be assigned to it. The property `hasCause` describes the cause of a causal relationship and has the causing node named "cause_node_name" as its value. Additionally, `hasEffect` describes the effect of the relationship and references the affected node named "effect_node_name". However, in the background, the creation of the CausalEdge also results in property assignments to the two affected CausalNodes. The edge with the outgoing edge gets the property `isCausing` with the edge object as the value. Thus, it is described that this node causes the other one via the affected edge. The affected node "effect_node_name" receives in return the property `isAffectedBy`, also specifying the affected CausalEdge.

> Listing 3: This code demonstrates the usage of `add.causal_edge()` to create `CausalEdges` between the two nodes "Rain" and "Wet", and give the edge the name "Rain->Wet". It's also possible to create an edge between two nodes that don't in the graph yet, "Wet" and "Slippery". By setting the `force_create` parameter to True, the method will automatically create the unknown nodes before creating the edge between them.

```
>>> # Create edge between existing nodes
>>> G.add.causal_edge(
...     cause_node_name="Rain",
...     effect_node_name="Wet",
...     name_for_edge="Rain->Wet")
>>> # Create edge between nodes and create unknown nodes
>>> G.add.causal_edge("Wet", "Slippery", "Wet->Slippery", force_create=True)
>>> # Printing list of graph individuals
>>> print(list(G.store.individuals()))

[cg_store.Rain, cg_store.Wet, cg_store.Rain->Wet, cg_store.Slippery, cg_store.Wet
    ->Slippery]
```

As can be seen in listing 4, the assignment of egde properties is possible. So there is the possibility to add a `confidence` in the range (0,1] or a `time_lag_s` in seconds to an edge, which could be useful for later evaluation by means of causal functions like causal discovery. Comments as free text are covered by the `comment` parameter.

> Listing 4: This example illustrates the possibility of assigning additional properties to edges, such as comments, confidences, or time lags. This allows for a more detailed representation of the causal relationships in the graph and can provide valuable additional information for analysis.

```
>>> G.add.causal_edge("Rain", "Wet", "Rain->Wet",
...     confidence=0.9,
...     time_lag_s=2.0,
...     comment=["some text"])
```

If created nodes are to be deleted again, this can be done via the variable `remove` and its function `causal_node()`. As with adding nodes, only the node name must be passed as a string for this, as shown in listing 5.

Listing 5: In order to remove an node, one can utilise the function `remove.causal_node()`. All that is necessary to do is to specify the name of the node, and the return value is either True or False, depending on the outcome of the execution.

```
>>> G.remove.causal_node("Rain")
```

In addition to the direct removal of individual edges via `causal_edge_by_name()`, under the specification of the edge name, two additional functions are available in `causalgraph`. So it is possible to remove all edges between two nodes. For this, the function `causal_edge()` must be passed the affected node names. The function `causal_edge_from_node()` can be used to remove all incoming and outgoing edges of this particular node by specifying a node name. For this purpose, only the name of the affected node must be passed to the function. Listing 6 shows the usage of those functions. The removal of edges directly influences the properties of the affected nodes. The edge objects are removed as values within the properties `isCausing` and `isAffectedBy`.

Listing 6: A demonstration of how the `remove` functions can be utilized in order to eliminate causal edges from a graph. The first approach to this process entails the removal of an edge by passing its name as a string. Alternatively, the second technique permits the removal of an edge by explicitly declaring the two nodes it links. Finally, the third line is instrumental in the removal of all edges connected to a specific node. This can be especially helpful in tidying up the graph and ridding it of unnecessary connections.

```
>>> # Remove edge by name
>>> G.remove.causal_edge_by_name("Rain->Wet")
>>> # Remove edge between two nodes
>>> G.remove.causal_edge("Rain", "Wet")
>>> # Remove all edges connected to specific node
>>> G.remove.causal_edge_from_node("Rain")
```

## 3.2 Importing and Exporting

Since `causalgraph` has made it its mission to act as an interface within the causal disciplines, the package offers import and export functionalities. These can be used to convert a `Graph()` object into the formats of other known packages of the adjacent causal methods or to transfer these formats into `causalgraph` again. These packages cover `Tigramite` [19], a package known for its causal discovery implementations, but also `NetworkX` [11], a commonly used Python package for creating and manipulating complex networks. In `causalgraph`, import and export functions are accessible through the `Load()` and `Export()` classes that are housed within the `Graph()` object.

Listing 7: Exporting graph G to `NetworkX` or `Tigramite` facilitates the utilization of tools and methods provided by these packages. For instance, taking advantage of causal discovery in `Tigramite`, it is possible to adjust the edges of a graph according to existing data sets. This example illustrates the usage of both `export()` methods.

```
>>> # Exporting Graph() to networkX.MultiDiGraph
>>> G_nx = G.export.nx()
>>> # Exporting Graph() to Tigramite Tuple
>>> G_tigra = G.export.tigra()
>>> # Alternate G_nx or G_tigra with other tools and methods
>>> # ... Do some_third_party_func(G_nx)
```

The `Export()` class provides multiple options for converting `Graph()` objects into other formats. One of these is the `export.nx()` function, which enables the conversion of a graph into a `NetworkX` `MultiDiGraph`. To avoid data loss and facilitate subsequent re-importing, all graph properties are stored as `NetworkX` attributes associated with the nodes and edges of the `MultiDiGraph`. These include comments, creator information, and properties from previously imported external ontologies. When exporting to a format that is compatible with `Tigramite` via

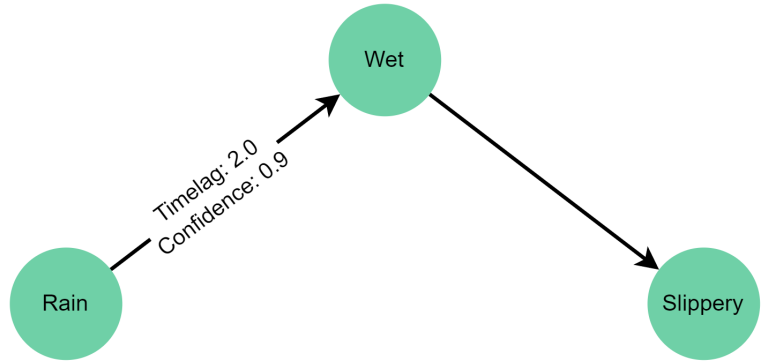

Figure 3: Result of applying the `draw.nx()` function from the proposed python package. The figure shows the causal relationship between rain and slippery ground, where the edges primarily represent the relationships and their properties, such as confidence and time delay. This visual representation facilitates the intuitive comprehension of these relationships.

`export.tigra()`, any additional properties are not retained. The resultant tuple object consists only of all information regarding simple nodes and their relations through edges, including the associated confidences, time lags, and the time step size of the time lags (expressed in seconds). All other data is not preserved. An alternative approach is to utilize the `export.gml()` or `export.graphml()` functions of the `causalgraph` package, which produce .gml or .graphml files respectively. These are both common formats that are supported by many graph-based packages. This makes it possible to transfer graphs into packages like DoWhy, utilize certain causal methods, and then transfer the results back into `causalgraph`. An example of this use case can be found in the fifth notebook of the proposed package examples[3]. An illustrative application of both functions can be observed in listing 7. The `Load()` class can be utilized for re-importing prior exported graphs. To this end, the function `load.nx()` is available for the import of `MultiDiGraphs`, and `load.tigra()` for the import of `Tigramite` tuples. It should be noted that this does not modify or fill the current `Graph()` object, but rather creates a new one. The update functions are intended to be included in subsequent releases. As demonstrated in listing 8, both of the above functions only require the external graph format to be passed to create a new `causalgraph` `Graph()` object. An already existing graph must be instantiated prior to executing the procedure. Nevertheless, `causalgraph` also offers the capability to fill graphs with nodes and edges from external graph formats when they are instantiated. To do this, the optional parameter `external_graph` must be passed to the `Graph()` constructor. This methodology can be observed in listing 8.

Listing 8: It is possible to generate a `Graph()` with the `load.nx()` function, which enables the user to utilize an existing graph. Furthermore, when utilizing the graph constructor, there is the option of providing an `external_graph` parameter at initiation, which allows for the graph to be filled with nodes and edges from other graph formats.

```
>>> # Create a second Graph() object by importing a MultiDiGraph
>>> G_nx_imported = G.load.nx(
...     nx_graph=G_nx,
...     sql_db_filnename="<path_to_new_sql_file>")
>>> # Load MultiDiGraph at Graph() init
>>> G_new = Graph(external_graph=G_nx)
```

### 3.3 Visualizing

The cause-effect relationships between different entities can be easily comprehended by humans when represented visually in the form of a graph. For this reason, `causalgraph` offers the ability to generate graphical representations of modeled `Graph()` objects. The `Draw()` class provides two main functions for this purpose: `draw.nx()` and

---
[3]All causalgraph example notebooks at hand at `https://github.com/causalgraph/causalgraph/tree/main/examples`

`html()`. The latter is based upon the graph visualization library `vis-network` [20] and generates a .html file that displays the graph with interactive capabilities. As depicted in Figure 3, `draw.nx()` produces a static image of the graph with properties such as time lag and confidence attached to the edges, if available. Additional information can be unveiled by hovering over the individual nodes. Both functions take the destination and the desired filename as parameters, where these are optional for draw.nx(). If omitted, the image is plotted directly using the Python package Matplotlib [21].

### 3.4 Third Party Ontologies

As briefly mentioned in chapter 2.3, ontologies can be extended with additional knowledge. In `causalgraph`, external knowledge in the form of additional ontologies can be imported and used. Listing 9 shows the import of the widely-used pizza ontology. It is sufficient to pass a URL of the third-party ontology to the `import_ontology()` function to make the classes of this ontology available for future reference. It is also possible to import local files. For this, the specific path to the ontology has to be passed. As can be seen in listing 9, individuals of newly imported classes are initially assigned only to their original type. Only when referenced in a `CausalEdge` they are inherited from the `CausalNode` class. After receipt of their `CausalNode` state, they are now also represented in the plotted graph. As can be seen in the listing, multiple inheritances are possible. Thus, a pizza continues to be a pizza even if it becomes a `CausalNode` by referencing it in a `CausalEdge`. This offers the advantage of utilizing all knowledge about a pizza, as well as being able to utilize the pizza in causal contexts. Thus, the example shows that a pizza Margherita is the cause of joy. The enrichment of metadata remains. However, additional ontologies can also be imported directly when initiating the `Graph()` object. For this, a list with ontology paths or URLs has to be passed as parameter `external_ontos` to the graph constructor.

Listing 9: Extension of the causalgraph ontology by the well-known Pizza ontology via the `import_ontology()` function. It is possible to import local and remote ontologies, for this the specific path to the ontology has to be passed. Once complete, the classes of the Pizza ontology are now available along with the `causalgraph` classes. Furthermore, a pizza inheriting the `CausalNode` class will retain its Pizza status at the same time.

```
>>> # Load third party ontology
>>> pizza_onto_url = "https://protege.stanford.edu/ontologies/pizza/pizza.owl"
>>> G.import_ontology(onto_file_path=pizza_onto_url)
>>> print(list(G.store.classes()))

[causalgraph.CausalEdge, causalgraph.CausalNode, ... pizza.PizzaBase, pizza.Food,
    pizza.ArtichokeTopping, pizza.VegetableTopping, pizza.Margherita]

>>> import causalgraph.utils.owlready2_utils as owlutils
>>> # Creating pizza object
>>> G.add.individual_of_type(class_of_individual="Margherita", "margherita_name")
>>> pizza_entity = owlutils.get_entity_by_name("margherita_name", G.store)
>>> print(pizza_entity.is_a)

[pizza.Margherita]

>>> G.add.causal_edge("my_margharita", "happiness", "my_edge", force_create=True)
>>> print(pizza_entity.is_a)

[pizza.Margherita, causalgraph.CausalNode]
```

## 4 Conclusion

The proposed Python package, `causalgraph`, is a valuable resource for researchers and practitioners in the field of causal discovery and causal inference. Developed by the Fraunhofer Institute for Machine and Tools and Forming Technology, the package is designed to model and store causal graphs embedded within knowledge graphs, serving as an interface for other tools in the causal discipline. Its main advantage is its ability to link additional information and metadata to causal structures, thereby preventing the loss of information and promoting seamless transfer between CD and CI.

The package includes several functions for creating and saving causal graphs, as well as for exporting the graphs for use in other packages. It also allows for the visualization of the graphs, which can help users to understand and interpret the data and relationships within the graph without further effort. This can significantly aid researchers and practitioners in their work, enabling them to work more efficiently with graph-based knowledge representations.

The development of `causalgraph` was motivated by the need to address the challenges posed by information loss and difficulty transferring data between CI and CD. This problem has been a significant obstacle for researchers in these fields, making it difficult to connect different causal methods effectively. The proposed package offers a solution to this problem, preventing information loss and facilitating smooth transfer between CD and CI.

Regarding potential uses and application scenarios, the package could be utilized in various settings, including research labs, academic institutions, and industry. It could also be used to support the development of new algorithms and models that rely on data from or for CD and CI. In the future, `causalgraph` could be expanded to support additional graph types and formats and incorporate new features and capabilities, such as partially updating or merging modeled graphs with imported graphs from third-party packages. As the field of CD and CI continues to evolve, the package could be updated to reflect these changes and to support the latest research and applications.

Overall, the development of `causalgraph` has the potential to significantly benefit researchers in the field of CD, CI, or adjacent methods, facilitating further progress in the area. The package will continue to be an essential tool for advancing the field in the future. As such, `causalgraph` is a valuable resource for researchers and practitioners working in this area, which should be encouraged to use in future research and applications.

## Acknowledgments

The development of `causalgraph` was part of the research project KausaLAssist. It is funded by the German Federal Ministry of Education and Research (BMBF) within the "Future of Value Creation - Research on Production, Services and Work" program (funding number 02P20A150) and managed by the Project Management Agency Karlsruhe (PTKA). The authors are responsible for the content of this publication.causa